\documentclass{article}

\usepackage{PRIMEarxiv}

\usepackage{framed,multirow}

\usepackage{amssymb}

\usepackage{url}
\usepackage[table,xcdraw]{xcolor}
\definecolor{newcolor}{rgb}{.8,.349,.1}

\usepackage{natbib}
\usepackage{hyperref}
\usepackage[switch,pagewise]{lineno} 
\usepackage{subcaption}
\usepackage{xspace}
\usepackage{url}
\usepackage{multirow}
\usepackage[justification=centering]{caption}
\usepackage{booktabs}
\usepackage{comment}
\usepackage[utf8]{inputenc}
\usepackage{lscape}
\usepackage{graphicx}
\usepackage{subcaption}

\newcommand{\ie}{\emph{i.e.,}\xspace}
\newcommand{\eg}{\emph{e.g.,}\xspace}
\newcommand{\etal}{\emph{et al.}\xspace}

\newcommand{\model}{RoBIn\xspace}
\newcommand{\genrobin}{RoBIn\textsuperscript{Gen}\xspace}
\newcommand{\extrobin}{RoBIn\textsuperscript{Ext}\xspace}
\newcommand{\red}[1]{{\color{red}{#1}}}

\usepackage{soul}

\graphicspath{{./figs/}{../}}

\title{RoBIn: A Transformer-based Model for Risk of Bias Inference with Machine Reading Comprehension
\thanks{\textit{\underline{Citation}}: 
\textbf{Authors. Title. Pages.... DOI:000000/11111.}} 
}

\author{
  Abel Corrêa Dias, Viviane Pereira Moreira, João Luiz Dihl Comba \\
  Instituto de Informática \\
  Universidade Federal do Rio Grande do Sul \\
  \texttt{\{abel.correa, viviane, comba\}email@inf.ufrgs.br} \\
}

\begin{document}
\maketitle

\begin{abstract}
\textit{Objective:} 
Scientific publications play a crucial role in uncovering insights, testing novel drugs, and shaping healthcare policies. Accessing the quality of publications requires evaluating their Risk of Bias (RoB), a process typically conducted by human reviewers. In this study, we introduce a new dataset for machine reading comprehension and RoB assessment and present \model (Risk of Bias Inference), an innovative model crafted to automate such evaluation. The model employs a dual-task approach, extracting evidence from a given context and assessing the RoB based on the gathered evidence.

\textit{Methods:} 
We use data from the Cochrane Database of Systematic Reviews (CDSR) as ground truth to label open-access clinical trial publications from PubMed. This process enabled us to develop training and test datasets specifically for machine reading comprehension and RoB inference. Additionally, we created extractive (\extrobin) and generative (\genrobin)  Transformer-based approaches to extract relevant evidence and classify the RoB effectively.

\textit{Results:} 
\model is evaluated across various settings and benchmarked against state-of-the-art methods for RoB inference, including large language models in multiple scenarios. In most cases, the best-performing \model variant surpasses traditional machine learning and LLM-based approaches, achieving an ROC AUC of $0.83$.

\textit{Conclusion:}
Based on the evidence extracted from clinical trial reports, \model performs a binary classification to decide whether the trial is at a \textit{low} RoB or a \textit{high}/\textit{unclear} RoB. We found that both \genrobin and \extrobin are robust and have the best results in many settings.

\end{abstract}

\keywords{Evidence-based Medicine \and Evidence-based Medicine \and Risk of Bias \and Tranformer-based Models \and Natural Language Processing \and Machine Reading Comprehension}

\section{Introduction}
\label{sec:introduction}
Randomized Clinical Trials (RCTs) are considered the gold standard for evidence in the biomedical field~\cite{Hariton2018-wt}. However, the way in which studies are conducted and results are published can sometimes introduce bias~\cite{Chalmers1981-tt}. The COVID-19 pandemic, which triggered a surge in biomedical research globally, exemplifies this issue. With an overwhelming number of publications emerging in a short time, the urgency to deliver rapid findings increased the risk of inaccuracies, misconduct, or retractions due to rushed peer reviews. Moreover, the pressure to produce immediate results during the pandemic created a fertile ground for commercial or political agendas to influence the interpretation of data, potentially leading to biased or skewed conclusions~\cite{Brainard2020}.

The Risk of Bias (RoB) is an important mechanism for evaluating the reliability of clinical trials and identifying any systematic errors that could occur during the planning or analysis phases~\cite{Berger&Alperson2009}. 
Misleading evidence distorts the true state of knowledge, making it difficult to recognize ignorance, unlike the clear acknowledgment of ignorance caused by a lack of evidence. Additionally, the presence of any clinical trial, even if it has the potential to mislead, hinders the start of new trials to explore the same questions. This underscores the importance of accurately assessing the RoB to protect the foundation of future biomedical research.

Several proposals defined instruments\footnote{These instruments are sometimes called \textit{tools}. However, we prefer the term \textit{instruments} to avoid confusing with computational tools.} 
for identifying sources of bias in clinical trials~\cite{Jadad+1996, Verhagen+1998, Moher+2003, Higginsd5928, Higgins+2019}. Despite the methodological differences between the instruments, they are related to the different ways in which bias might be introduced into the result (bias type or bias domain). For example, the Cochrane RoB Tool~\cite{Higgins+2019} comprises guidelines based on signaling questions (\ie \textit{yes, no} or \textit{not informed} questions) to be answered by a human reviewer, followed by the manual extraction of evidence sentences that support the answer. After answering these questions, reviewers assess the RoB for the entire document. 

The RoB assessment in clinical trials was approached in the literature using various machine learning models. Examples include Support Vector Machines (SVMs)~\cite{Marshall+2015, Marshall+2016, Pereira+2020}, Convolutional Neural Networks (CNNs)~\cite{Zhang+2016}, or Logistic Regression (LR)~\cite{Millard+2015, Marshall+2020}.
More recently, Large Language Models (LLMs) were employed to determine the RoB in RCTs~\cite{Lai+2024}.
Although these studies developed RoB datasets to train and test their methods, none have publicly shared these datasets, making comparison challenging or even unfeasible. 

This work introduces the Risk of Bias Inference (\model) dataset, which, to the best of our knowledge, is the first publicly available dataset focused on RoB assessment. The \model dataset was constructed using distant supervision, where existing databases are utilized to label the data~\cite{Mintz+2009, Nguyen&Moschitti2011}. 
In this setup, each tuple consists of a question-answer pair, where the question is a signaling question and the answer is supported by evidence, contributing to the final RoB judgment. The class label indicates the answer to the signaling question, determining whether the RoB is classified as \textit{low} or \textit{high/unclear}.

Transformers~\cite{Vaswani+2017} are the state-of-the-art in language processing tasks and have been successfully employed in many biomedical applications~\cite{Madan2024}, including argument mining~\cite{STYLIANOU2021103767}, text mining~\cite{Lee+2019}, text analysis~\cite{CAI2023104418}, information retrieval~\cite{LOKKER2023104384}, among others. 
Transformer‑based models have also been trained and deployed  
for analyzing biomedical‑related datasets~\cite{Madan2024}.
Using the \model dataset, we trained Transformer-based models to extract supporting sentences and perform RoB assessments. We implemented both extractive (\extrobin) and generative (\genrobin) approaches, comparing their performance with commonly used baselines. 

Our experimental evaluation demonstrated that both \extrobin and \genrobin performed well on the MRC task across all bias types and achieved competitive results for RoB classification, even outperforming classical methods and LLMs.

\section{Background}
\label{sec:background}
This section introduces the definition of clinical trial quality and the RoB, describing the most common instruments for clinical trial assessment. Next, we introduce NLP tasks that are close to our research problem, and finally, we review the Transformer architecture for NLP tasks.

\subsection{Clinical Trial Quality}
\label{subsec:background-trial-quality}

Medical research can be classified into primary and secondary research. Primary research is divided into three main areas: basic, clinical, and epidemiological. These research areas aim to make discoveries and provide clear evidence of the efficacy or safety of drugs or medical devices~\cite{Rohrig+2009}. The secondary research is divided into meta-analyses and reviews. A meta-analysis summarizes quantitative data about primary research studies using statistical methods.
Conversely, the reviews are qualitative summaries that can contain meta-analyses. A narrative review provides a broad overview of the topic, while a systematic review aims to analyze all the literature around a research question~\cite{Ressing+2009}. This work deals with two types of clinical studies: clinical trials and systematic reviews. The assessment of the RoB, also known as critical appraisal or quality assessment, occurs in the RoB stage of the systematic review. It aims to screen and exclude studies from the systematic review based on relevance and evaluation of the RoB~\cite{Tsafnat+2014, Frampton+2022}.

The RoB assessment corresponds to a validity check of clinical studies 
to screen and exclude studies from the systematic review based on relevance and RoB evaluation~\cite{Tsafnat+2014, Frampton+2022}. Although more than 170 non-redundant concepts are related to different bias types~\cite{ALPER2021103685},
in practice, only the five more important bias types are considered when interpreting clinical trials~\cite{Phillips+2022}: \textbf{Selection bias} is due to methods used to assign patients to study treatment groups; \textbf{performance bias} occurs when some of the participants or the staff are aware of the assigned treatment; \textbf{detection bias} occurs in the measurement of the outcomes when assessors are aware of the assigned treatment; \textbf{attrition bias} occurs as a result of patient withdrawals that affect a certain subset of the patients, and \textbf{reporting bias} may occur when non-significant findings are ignored or omitted from the results.

Existing instruments for assessing the RoB in RCTs involve a set of guidelines, methodologies, or procedures. For example, 
the \textbf{Jadad score}~\cite{Jadad+1996}, is based on a scoring system ranging from $0$ to $5$ focused on three \emph{yes/no} questions concerning the randomization process, blinding, withdrawals, and dropouts of RCTs. The \textbf{Delphi List}~\cite{Verhagen+1998} defines a set of nine \emph{yes/no/don't know} questions to assess quality in RCTs. A group of experts agreed on this number of questions after voting on questionnaires with lists of criteria. In addition to the randomization and blinding processes, the Delphi List includes questions about the allocation of patients, eligibility criteria, statistical analysis, and intention-to-treat analysis. 
The \textbf{CONSORT statement}~\cite{Altman+2001, Moher+2003} contains a checklist of essential items that should be included in an RCT and a flow diagram for correctly reporting results. It was not designed to be a quality assessment tool. The \textbf{Revised Cochrane Risk of Bias tool for randomized trials} (RoB 2)~\cite{Higgins+2019} is a framework for assessing the RoB in five bias types\footnote{we use the term \textit{bias type} to refer to the \emph{bias domain} used in RoB 2.}.
RoB 2 contains signaling questions for each bias type, with 5-scale option answers: \textit{yes}, \textit{probably yes}, \textit{no}, \textit{probably no}, and \textit{no information}, and free-text boxes to include additional information or quotes from the text to provide support. The RoB 2 algorithm maps the answer to the signaling question to three possible judgments: \textit{low RoB}, \textit{some concerns}, and \textit{high RoB}. Despite the algorithm output, the reviewer can double-check and change if appropriate.

It should be acknowledged that researchers do not universally agree on the exact meaning of the word \textit{quality}. As a result, the notion of quality remains undefined and open to interpretation. Usually, it is associated with evaluating the risk of biased results in the studies included in systematic reviews. 

\subsection{Machine Reading Comprehension}
\label{subsec:background-mrc}
Machine reading comprehension (MRC) is a challenging task that aims at creating systems that can answer questions given a context. Instances in an MRC dataset comprise a set of contexts denoted $\mathcal{C}$, questions $\mathcal{Q}$, and answers $\mathcal{A}$. For a given question $q_i \in \mathcal{Q}$, and its corresponding context $c_i \in \mathcal{C}$, the goal is to learn a function $f$ $ f: (\mathcal{C}, \mathcal{Q}) \rightarrow \mathcal{A}$ which extracts/generates the answer $a_i \in \mathcal{A}$, by receiving as input the question and the context.

The main difference between question-answering (QA) and MRC is that in QA, the only input is the question, while in MRC, the inputs are the question and the corresponding context from which the answer should be extracted~\cite{Baradaran+2022}. The context can consist of one or more sentences.
The questions can be grouped into three categories: factoid questions, non-factoid questions, and yes/no questions~\cite{Baradaran+2022}. Factoid questions can be answered in terms of named entities or short text answers. On the other hand, non-factoid questions usually require long and complex answers composed of more than one passage. 
In this article, we focus only on yes/no questions. 

Regarding the approach employed by the MRC system, the output can be classified into \textit{generative} or \textit{extractive}. In the extractive approach, the answer is a span from the context, while in the generative approach, the answer is generated according to the context and question. 

\subsection{Transformer Architecture}
\label{sec:transformer}
The Transformer architecture~\cite{Vaswani+2017} has revolutionized many NLP tasks due to its capacity to model contextual dependencies within tokens and sentences. It introduced novel design principles to address the limitations of CNNs and recurrent neural networks (RNNs). Unlike models that process text sequentially, Transformers process text in parallel, capturing both left and right contexts for each word. They employ self-attention mechanisms to process input data in parallel. The parallelization and, consequently, scalability of the Transformer architecture make it the foundation for state-of-the-art models in many tasks, including text classification, machine translation, and text generation. Subsequent works have extended the Transformer architecture, yielding more remarkable advancements in NLP tasks~\cite{Radford+2018, Devlin+2019, Lee+2019, Gururangan+2020}.

Transformer models aim to learn sentence representations that can be easily adapted to many downstream tasks by combining unsupervised (or semi-supervised) pre-training and supervised fine-tuning strategies. The most common approach is to apply a language modeling objective on unlabeled data to learn the initial parameters. Next, the parameters are adapted using the corresponding task objective and labeled data.

\section{Related Work}
\label{sec:related-work}
 
The first attempt to automate RoB was made in 2014~\cite{Marshall+2014}. A scoping review identified $11$ studies regarding the analysis of RoB~\cite{Santos+2023}. Most of these studies report the use and effectiveness of existing methods rather than proposing new solutions. Currently, no tool can perform all tasks, from creating to publishing a systematic review in an automated manner. 

The initial efforts for automatically assessing RoB employed soft-margin SVMs for sentence identification and document classification modeled as two separated binary classification tasks to predict the RoB for each bias type of the Cochrane RoB tool~\cite{Marshall+2015}. Next, a joint model was created to predict the RoB over all bias types and extract the supporting sentences~\cite{Marshall+2015, Marshall+2016}. Given the lack of machine-readable data for the RoB, authors extracted data from the Cochrane Database of Systematic Reviews (CDSR)~\footnote{Cochrane Database of Systematic Reviews: \url{https://www.cochranelibrary.com/cdsr/reviews}} using distant supervision to assign weak labels to a corpus of 2200 RCTs in PDF format. 

A family of CNN models was used to perform document-level classification by leveraging document labels and rationales provided by annotators~\cite{Zhang+2016}. The rationales are snippets of text that support the categorization of the document. 
The Rationale Augmented CNN (RA-CNN) was evaluated on a dataset with the following bias types: random sequence generation, allocation concealment, blinding of participants and personnel, and blinding of outcome assessment. RA-CNN outperformed baseline approaches (including SVM and other CNN-based methods) and provided better interpretability by highlighting the rationales that contributed most to the classification. 
However, the quality metrics failed to show significant differences compared to methods such as SVM. In addition, the CNN-based method also struggles to handle long-range contextual dependencies on sequential data.

LR models have also been used for sentence identification and RoB assessment for sequence generation, allocation concealment, and blinding~\cite{Millard+2015}. The sentence-level model ranks sentences by relevance, while the article-level model aims to rank articles by their RoB. 
%

TrialStreamer~\cite{Marshall+2020} is a system that continuously evaluates abstracts from RCT publications by extracting relevant information and providing an overall RoB score based on LR with L2 regularization, displaying the probability of a given RCT having a low RoB.
The evaluation metrics, however, have shown low scores in binary classification (\ie F1 = 0.45).  

Similarly, MCRB (Multi Classifier for Risk of Bias)~\cite{Pereira+2020} is a system to automate the RoB assessment in clinical trial reports across different bias types. Authors collected and preprocessed $237$ articles and applied traditional machine learning algorithms.
RoB results were $AUC = 0.77$ on average for all bias types, with the attrition bias having the lowest score  ($AUC = 0.45$).

A recent study showed promising results applying Large Language Models (LLMs) to assessing the RoB~\cite{Lai+2024}. Two LLMs (ChatGPT and Claude) were selected for the study. A multidisciplinary panel of evidence-based medicine and computer science experts developed prompts and oversaw the assessment process. Thirty randomly chosen RCTs were used to evaluate the LLMs according to the accuracy and consistency of the predictions under criteria defined by the experts. Their results demonstrated that LLMs can achieve a high accuracy and consistency compared to human reviewers in assessing RoB in RCTs. Some limitations of the work are the small sample size, the inability to consider supplementary materials, and lengthy appendices when assessing the RoB.
While the authors highlight the capabilities of LLMs in automating the assessment of RoB, there are some cautions against the uncritical adoption of these technologies due to significant risks and limitations. LLMs can produce confidently written outputs that are factually incorrect or misleading due to hallucinations or omission of critical details, leading to the dissemination of false medical information. It can be dangerous, particularly in healthcare settings. Other concerns are the lack of accountability for the outputs generated by LLMs, fabricated data such as references and statistics, misleading conclusions, and unclear provenance of the origins of the information used by LLMs~\cite{Yun+2023}.

Table~\ref{tab:comparison-related-work} summarizes all the approaches discussed in this section. Most rely on bag-of-word models and traditional machine learning algorithms such as SVM and LR. Another point is that most datasets were created from scratch using systematic reviews and clinical trial reports that are publicly available. However, those datasets are unavailable, which brings us to the need to develop our own using the same techniques already used in the existing studies (\ie distant supervision).

Our work differs from existing solutions in a number of ways.
To the best of our knowledge, this is the first work to implement a Transformer-based model to perform RoB assessment.
Unlike TrialStreamer~\cite{Marshall+2020}, we analyze the complete contents of the article describing the clinical trial and not just the abstract. Furthermore, besides showing the overall RoB assessment, we include results for each bias type.
In addition, some previous works employed a bag-of-words approach~\cite{Marshall+2015, Millard+2015, Marshall+2016}. The limitations of such an approach relate to the need for manual feature extraction and the fact that it disregards word order, which can lead to the loss of semantics. Unlike those, our work relies on contextual embeddings, which can better represent the semantics of the text.

Furthermore, most work on automating the assessment of RoB has been carried out by a single research group from King's College London~\cite{Marshall+2016, Marshall+2015, Marshall+2020, Zhang+2016}. In this sense, our work contributes by adding more diversity to the research area.

\begin{table}[tph]
\centering
\caption{Approaches for automating RoB assessment using binary classification. All datasets are not public.}
\label{tab:comparison-related-work}
\resizebox{0.95\columnwidth}{!}{%
\begin{tabular}{llp{5cm}p{5cm}}
\hline
\multicolumn{1}{c}{\textbf{Work}}                & \multicolumn{1}{c}{\textbf{ML Algorithm}}                                                 & \multicolumn{1}{c}{\textbf{Dataset}}                                                                                                             & \multicolumn{1}{c}{\textbf{Results}}                                                                                                                                                                                                              \\ \hline
Marshall~\etal\cite{Marshall+2015} & SVM                                                                                       & 2200 clinical trial reports automatically annotated                                                                              & \begin{tabular}[t]{@{}l@{}}
F1: 0.70\\
Precision: 0.64\\
Recall: 0.80\\\end{tabular}                                               \\ \hline
Millard~\etal~\cite{Millard+2015}                 & LR                                                                       & \begin{tabular}[t]{@{}p{5cm}@{}}1467 articles from PubMed with CDSR RoB tool assessments\end{tabular} & \begin{tabular}[t]{@{}p{5cm}@{}}Ranking articles by RoB:\\ (full-text): AUC $> 0.72$\\ (title and abstract): AUC $>0.67$\end{tabular} \\ \hline
Marshall~\etal~\cite{Marshall+2016}                & SVM                                                                                       & 12808 CDSR RCTs                                                                         & \begin{tabular}[t]{@{}l@{}}Accuracy for labeling documents \\with RoB: $0.71$\end{tabular}                                                                                                                                                               \\ \hline
Zhang~\etal~\cite{Zhang+2016}                   & CNN                                                                                       & 12808 CDSR RCTs - from~\cite{Marshall+2016}                                                                              & \begin{tabular}[t]{@{}p{5cm}@{}}Accuracy for labeling documents \\with RoB:\\ RA-CNN: $0.74$\\ Human reviewer: $0.81$\end{tabular}   \\ \hline
Marshall~\etal~\cite{Marshall+2020}                & LR                                                                       & 13463 CDSR RCTs with RoB tool assessments
& \begin{tabular}[t]{@{}p{5cm}@{}}F1: $0.45$\end{tabular}                                                        \\ \hline
Pereira~\etal~\cite{Pereira+2020}                 & \begin{tabular}[t]{@{}l@{}}LR\\ NB\\ SVM\\ XGB\end{tabular} & 246 articles with RoB assessments                                                                       & \begin{tabular}[t]{@{}l@{}} 
LR (F1: 0$.68$, AUC: $0.77$) \\
NB (F1: 0$.64$, AUC: $0.72$) \\
SVM (F1: 0$.69$, AUC: $0.77$) \\
XGB (F1: 0$.86$, AUC: $0.74$) \\
\end{tabular}                        \\ \hline
Lai~\etal~\cite{Lai+2024}                     & \begin{tabular}[t]{@{}l@{}}LLM ChatGPT \\ LLM Claude \end{tabular}                                                                                   & 30 randomly selected RCTs                                                                                           & \begin{tabular}[t]{@{}l@{}} ChatGPT (F1: $0.78$)\\ Claude (F1: $0.73$)\end{tabular}                                                                                                                                                          \\ \hline
\end{tabular}
}
\end{table}

\section{\model Dataset}
\label{sec:dataset}
Assessing the RoB for a clinical study is often performed manually. However, as mentioned in Section~\ref{sec:related-work}, previous work showed potential to automate this task using distant supervision and machine learning techniques~\cite{Marshall+2015, Marshall+2016}. Similarly, LLMs demonstrated promising results~\cite{Lai+2024}. Given the absence of public datasets for the specific task we want to address, our first step was constructing one. The \model dataset is publicly available\footnote{\url{https://github.com/phdabel/robin}}.

\begin{figure*}[t!]
    \centering
    \includegraphics[width=0.9\textwidth]{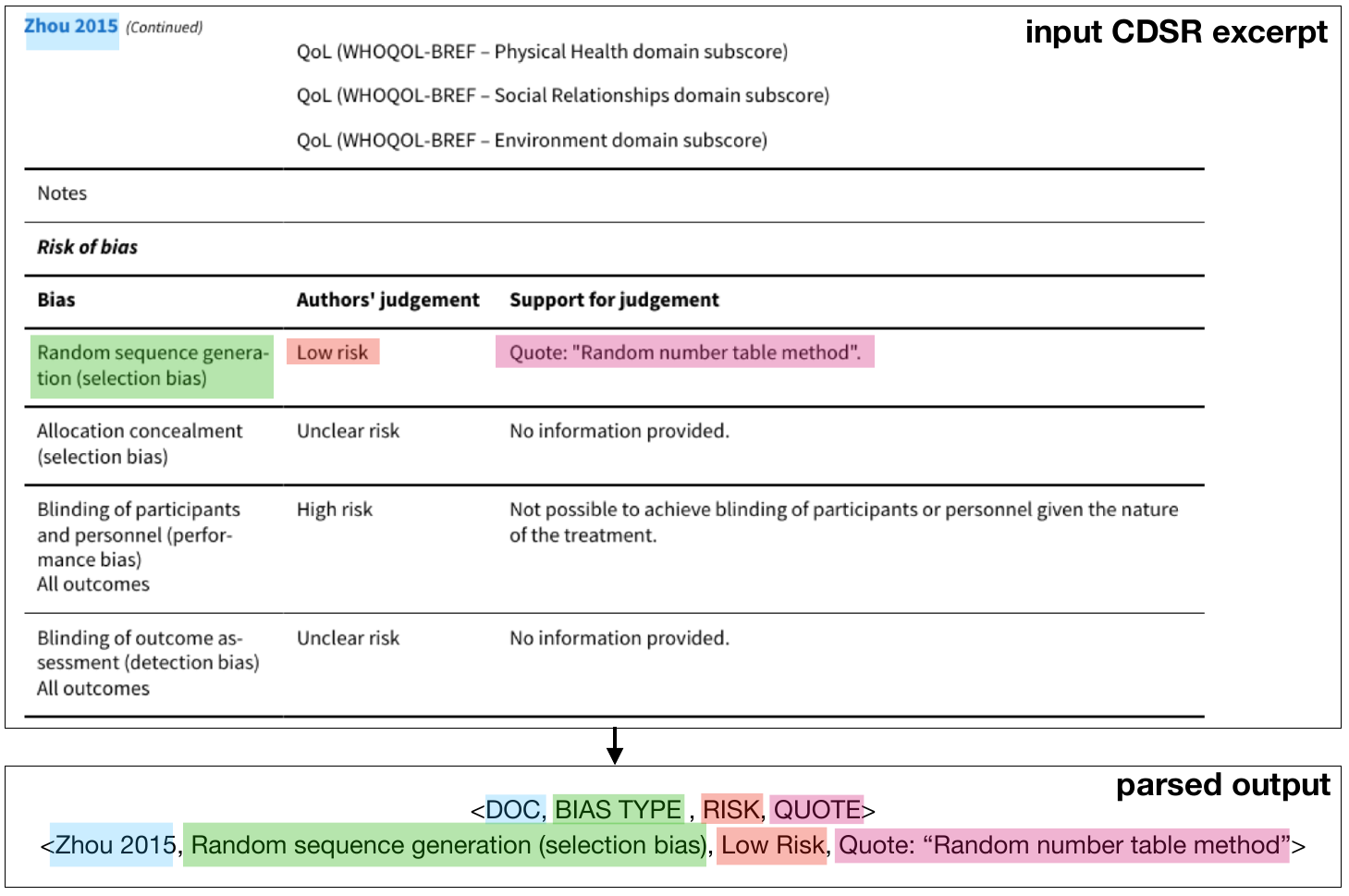}
    \caption{Parsing the systematic reviews to obtain instances for the dataset.}
    \label{fig:cdsr-data-coll}
\end{figure*}

CDSR is a comprehensive resource in healthcare encompassing systematic reviews, protocols, editorials, and supplementary materials. The systematic reviews offer insights into various studies through a detailed methodology. Furthermore, they include tables that provide details on references, evaluative judgments, and evidence supporting the judgment of multiple bias types (\eg supporting sentence). Figure~\ref{fig:cdsr-data-coll} illustrates the process of parsing a systematic review table into a structured document. For each table, we extract the reference of the study, the bias type, the RoB label, and the supporting sentences.
Additionally, reviewers compile a bibliography for each examined study question, yet the dataset lacks explicit linkage between each supporting sentence and its corresponding excerpt in the references. Also, there are cases in which the supporting sentences are comments from the reviewers. This absence of direct connections between references and supporting sentences might pose challenges for traceability and verification in the annotation process.

\subsection{Dataset components}
\label{sec:dataset-components}
The components of the proposed dataset are:

\begin{enumerate}
    \item Bias type -- selection, performance, detection, attrition, reporting, and other bias;
    \item Question -- the signaling question associated with the bias type;
    \item Context -- the set of sentences encompassing the supporting sentence;
    \item Supporting sentence -- from the systematic review author;
    \item Answer start -- marking the supporting sentence offset within the context;
    \item Ground truth class labels for RoB judgment (\ie, low, or high/unclear).
\end{enumerate}

For data provenance, we also keep additional information, such as codes used to identify the study and the systematic review within the CDSR, the PubMed ID, and the row used to determine where the context was found within the article. 

\subsection{Creating the dataset} 
\label{subsec:creating-dataset}
We acquired systematic reviews from the Cochrane Organization after signing a non-disclosure agreement that allowed us to use the data for research purposes. Each systematic review is in an RM5 file, a specially designed XML format to store systematic reviews. The systematic review dataset comprises structured data with tables containing RoB judgments and supporting evidence from human reviewers for different bias types. We used the extracted data as ground truth to automatically annotate the \model dataset.

\begin{figure}[!t]
    \centering
    \includegraphics[width=0.8\linewidth]{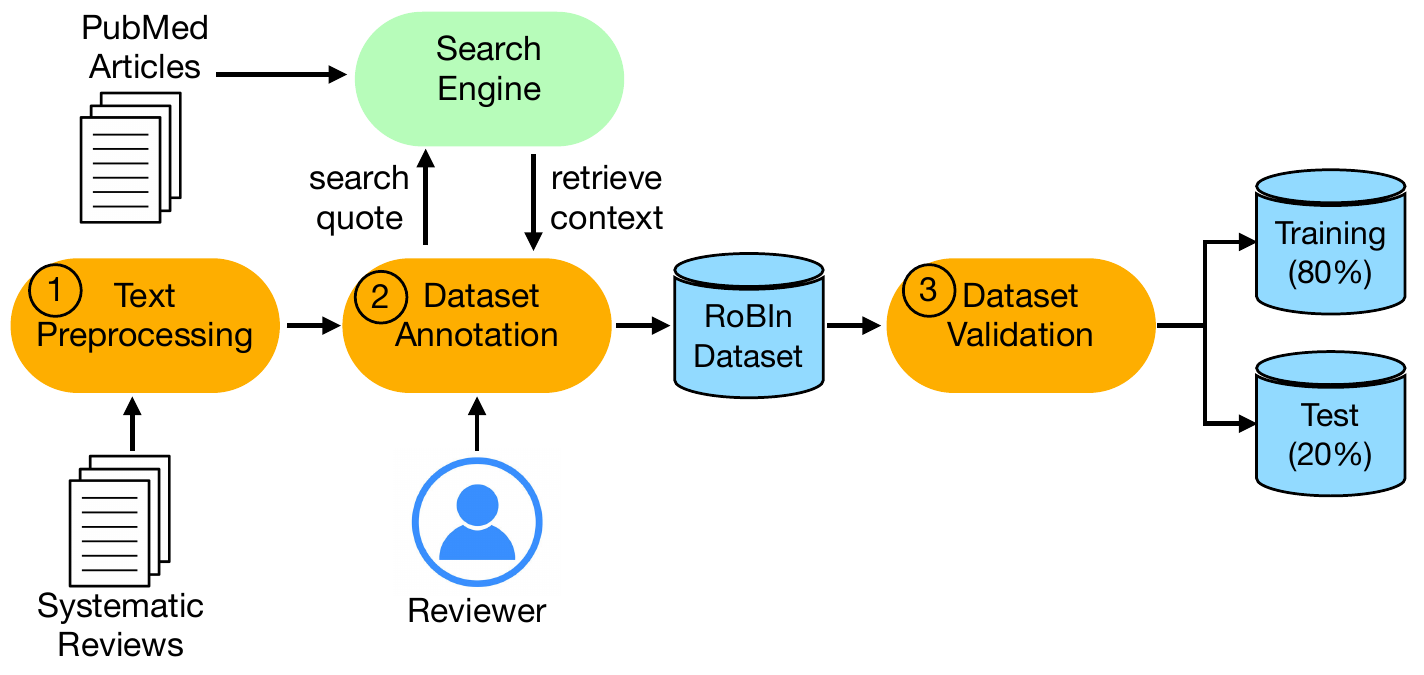}
    \caption{Steps for creating the \model dataset.}
    \label{fig:qa-robin-ds-pipeline}
\end{figure}

Figure~\ref{fig:qa-robin-ds-pipeline} illustrates the process we followed in the development of the \model dataset. The initial phase involved acquiring all publicly available articles from PubMed and PubMed Central. This extensive collection was filtered automatically to retain only those articles explicitly cited in the reference sections of the systematic reviews under consideration, with all non-referenced articles being excluded from the dataset. As a result, we obtained $7,334$ articles. 

Next, each article was segmented into sentences, and the sentences were indexed by a search engine.  The goal is to identify sentences cited as supporting evidence in the systematic reviews.

After indexing the articles, we began the process of labeling the dataset. Using the database obtained from Cochrane, we pre-processed the RoB tables to extract relevant information, such as the bias judgments and the supporting sentences (Step~1 in Figure~\ref{fig:qa-robin-ds-pipeline}). The annotation process (Step~2 in Figure~\ref{fig:qa-robin-ds-pipeline}) consists of searching the supporting sentences in the indexed articles to get the answer and the context. Usually, the supporting sentences are direct quotes from the articles, but in some cases, they are paraphrased or include comments from the reviewer. In those cases, linking the supporting sentences and the correct part is more difficult. Thus, we relied on the cosine similarity to compare the sentences retrieved from the articles with the extracted supporting sentences from the systematic reviews. Sentences with cosine similarity higher than $0.5$ were filtered, and the one with the highest similarity was selected.

Having a more comprehensive context, which includes the supporting sentence, is useful for dataset users to gain a better insight into the relevance and implications of each supporting sentence within the broader narrative of the article. 
To create the context, we extracted the sentences surrounding the supporting sentence, \ie we kept at most three sentences before and after it. By offering a contextualized view of each supporting sentence, \model dataset is a useful resource for training and evaluating MRC models for retrieving evidence to assess the RoB. 

During the validation process (Step~3 in Figure~\ref{fig:qa-robin-ds-pipeline}), one of the authors, 
curated a collection of $23,873$ instances, eliminating $6,915$ instances due to parsing errors or inconsistent results (\ie answers retrieved that do not correspond to the supporting sentences). We divided the remaining instances into training ($13,554$) and test sets ($3,404$). The stratification process was guided by the prevalence of each bias type in the dataset, ensuring that each was proportionally represented in the training and test set to mitigate any potential bias in model training and evaluation.

\begin{figure*}[t!]
    \centering
    \includegraphics[width=0.9\textwidth]{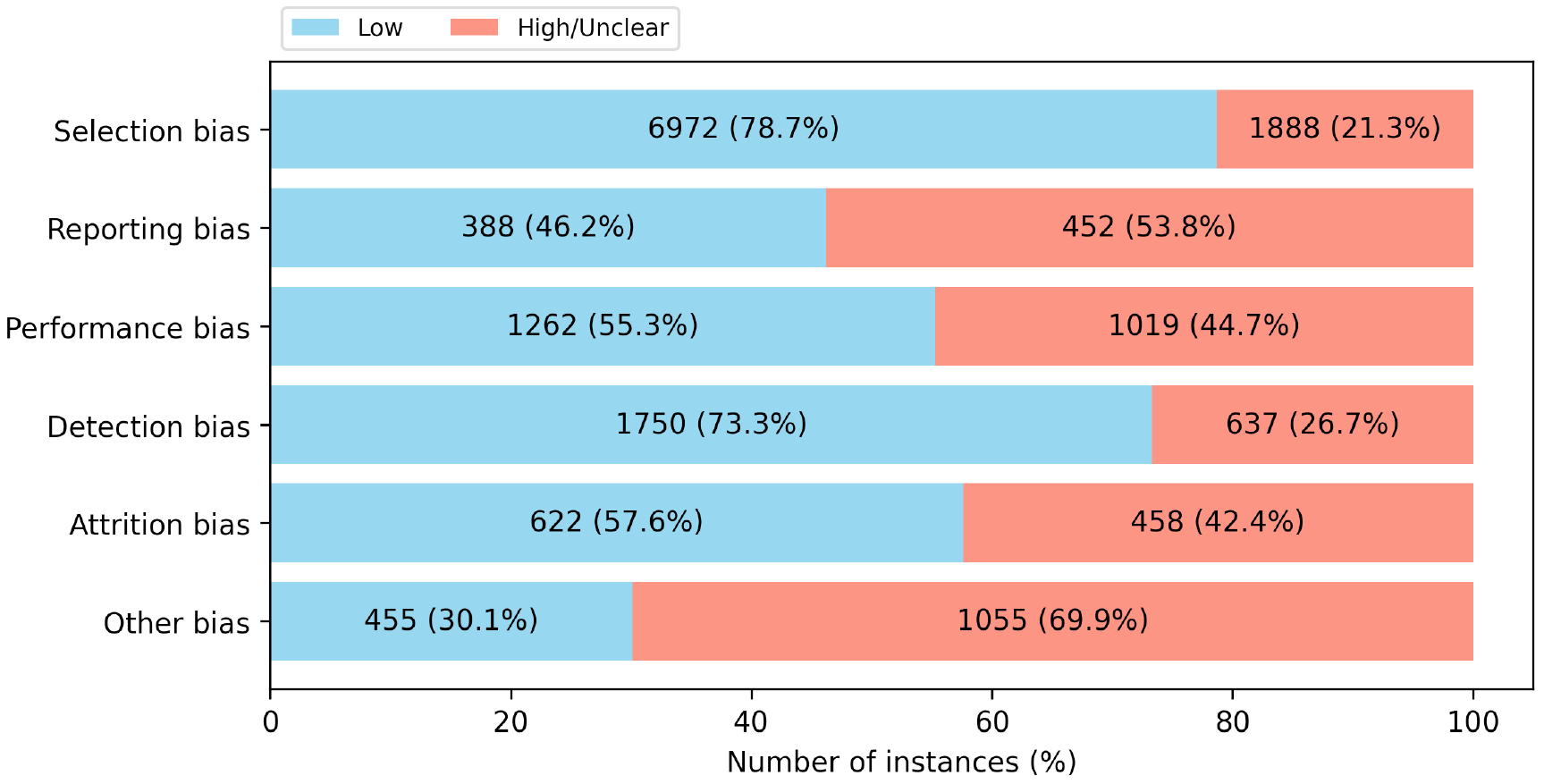}
    \caption{Class Distribution by bias type}
    \label{fig:dataset-statistics-by-type-bias}
\end{figure*}

For each bias type, our approach relies on the following signaling questions from the Cochrane methodology~\cite{Higgins+2019}: 
\begin{itemize}
\setlength\itemsep{0em}
    \item Was the allocation sequence random? (selection bias)
    \item Was the allocation sequence concealed until participants were enrolled and assigned to interventions? (selection bias)
    \item Were the trial participants, staff, and study personnel blind to the intervention? (performance bias)
    \item Was the outcome assessor blinded to the intervention? (detection bias)
    \item Was the incomplete outcome data addressed? (attrition bias)
    \item Is the study free from selective reporting? (reporting bias)
    \item Is the study free from other types of bias? (other bias)
\end{itemize} 

To add the signaling questions aligned with Cochrane's bias types to our dataset, we devised a simple heuristic. The goal was to link specific signaling questions to the corresponding bias types. This simple heuristic was designed based on the analysis of common patterns and terminologies observed within the dataset that resonated with the essence of the signaling questions in Cochrane's methodology.
The distribution of instances in \model dataset is illustrated in Figure~\ref{fig:dataset-statistics-by-type-bias}.

\section{\model Model}
\label{sec:methods}
In this section, we define the two problems we are addressing and introduce two Transformer-based models designed to retrieve the supporting sentences (the MRC problem) and classify the RoB from clinical trial texts (the RoB inference problem). We considered extractive (\extrobin) and generative (\genrobin) approaches. The extractive model is based on the encoder of the Transformer model~\cite{Vaswani+2017}, while the generative model leverages both the encoder and decoder of the Transformer. 

Both models consist of a MRC task (see Section~\ref{subsec:background-mrc}) followed by a RoB assessment binary classification, with the class labels \textit{yes} and \textit{no}/\textit{not informed} corresponding to \textit{low} and \textit{high}/\textit{unclear} RoB.
This dual functionality is especially useful in scenarios where extracting relevant information from a context is required and where a subsequent classification of the extracted information is necessary, such as determining the reliability or sentiment of the answer. For example, considering the bias related to the randomization process, simply stating that the process was carried out randomly without detailing some aspects of the process is not enough to assess the risk as low. Thus, a machine-learning model must learn when the answer is appropriate for judging the RoB as low.

We address RoB Inference as a classification task, where, given a set of answers $\mathcal{A} = \{ a_1, a_2, \cdots, a_n \}$, such that each answer is a sequence of words $a_i = \{ w_{i1}, w_{i2}, \cdots, w_{im} \}$, and a binary label $\mathcal{Y} = [ 0, 1 ]$, the goal is to find a classification function $g: \mathcal{A} \rightarrow \mathcal{Y}$ that maps each answer $a_i$ to a binary label $y_j$.

\begin{figure*}[ht!]
    \centering
    \begin{subfigure}[b]{0.2\linewidth}
        \centering        \includegraphics[width=\textwidth]{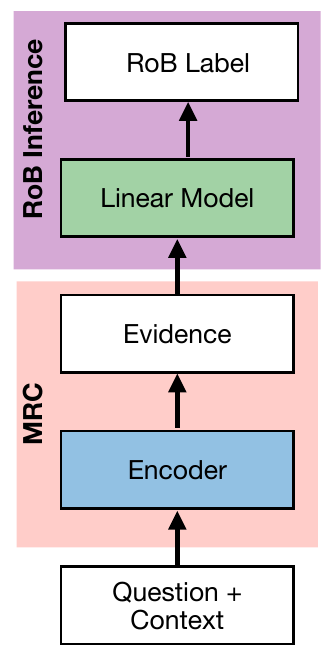}
        \caption{\extrobin}
        \label{fig:extractive-robin-model}
    \end{subfigure}
    \hspace{1cm}
    \begin{subfigure}[b]{0.2\textwidth}
        \centering        \includegraphics[width=\textwidth]{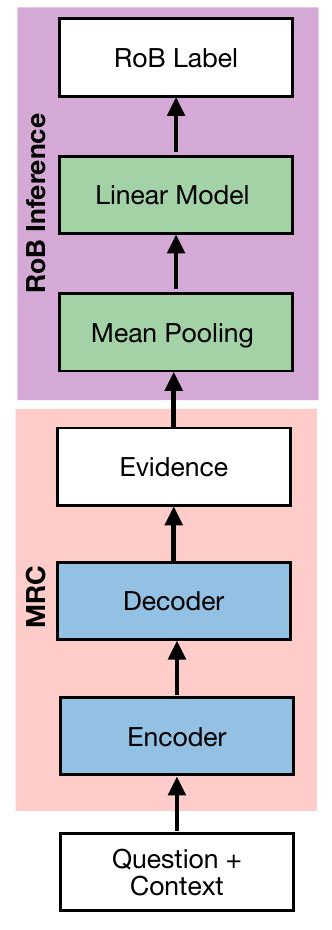}
        \caption{\genrobin}
        \label{fig:genrobin-model}
    \end{subfigure}    
    \caption{Architecture of the Extractive and Generative \model models}
    \label{fig:robin-architecture}
\end{figure*}

Figure~\ref{fig:extractive-robin-model} illustrates the architecture of \extrobin. The first steps correspond to the typical MRC setting in which an input question and context are given. Recall that our context is a set of contiguous sentences from the clinical trial report. The encoder extracts a piece of evidence from the context. The evidence is submitted to a classification model that outputs the RoB label.

In contrast, \genrobin (Figure~\ref{fig:genrobin-model}) involves both an encoder and a decoder. The question and context are passed to the encoder and the evidence is passed to the decoder. 
This combination allows us to model the probability of a class (\ie probability of low RoB) and also apply a threshold to perform binary classification.

\subsection{Extractive \model}
\label{sec:extractive-model}
For training \extrobin, we initialized the model with pre-trained parameters and fine-tuned it for the MRC task. We represent the question and the context as a single-packed sequence and trained two sets of parameters, a start vector $S \in \mathbb{R}^{|H|}$, and an end vector $E \in \mathbb{R}^{|H|}$, where $|H|$ is the hidden layer size~\cite{Devlin+2019}. For a given token, $\hat{y}_{s,i} = \frac{\exp^{S \cdot T_i}}{\sum_j \exp^{S \cdot T_j}}$ denotes the probability of $i$ being the start of the answer span, whereas $\hat{y}_{e,i} = \frac{\exp^{E \cdot T_i}}{\sum_j \exp^{E \cdot T_j}}$ denotes the probability of $i$ being the end of the answer, where $T_i \in \mathbb{R}^{|H|}$ is the hidden vector for the input tokens. The score of the candidate span is used for prediction. The training objective, denoted as $L_{MRC}$ in Equation~\ref{eq:qa-loss}, is the weighted sum of the log-likelihoods for the correct start and end positions (\textit{i.e.}, $y_{s,i}$, and $y_{e,i}$, respectively).

\begin{equation}
    \label{eq:qa-loss}    
    L_{MRC} = - \frac{1}{2} \left( \displaystyle\sum_{i=1}^N y_{s,i} \log(\hat{y}_{s,i}) + \displaystyle\sum_{i=1}^N y_{e,i} \log(\hat{y}_{e,i}) \right)
\end{equation}

The classification task is achieved by training separated models with the expected evidence. We employed the cross-entropy loss, denoted as $L_{CE}$ in Equation~\ref{eq:ce-loss}, which measures the difference between the distributions of the true labels, denoted as $y_{i,c}$, and the labels predicted by the model, denoted as $\hat{y}_{i,c}$.

\begin{equation}
    \label{eq:ce-loss}
    L_{CE} = -\frac{1}{N}\displaystyle\sum_{i=1}^N \displaystyle\sum_{c=1}^C y_{i,c} \log \left( \frac{e^{\hat{y}_{i,c}}}{\sum_{k=1}^C e^{\hat{y}_{i,k}}} \right)
\end{equation}

After training the classification models, we use them to infer the RoB on the evidence extracted using \extrobin. This stage aims to classify the extracted answer into a high/unclear or low RoB.

\subsection{Generative \model}
\label{sec:generative-model}
We also initialized the \genrobin with pre-trained parameters and fine-tuned it for the sequence generation task. We concatenate the question and the context, passing them as input to the encoder, while the expected answer is passed as input to the decoder. The score of the decoder output is used for prediction.

In sequence-to-sequence models, the cross-entropy loss is computed token-by-token through the output sequence. The training objective, denoted as $L_{G}$ in Equation~\ref{eq:loss-generative}, represents the cross-entropy loss for a prediction $\hat{y}$ and the actual sequence $y$, where $T$ is the length of the sequence, $y_t$ is the token at the position $t$, and $P(.)$ represents the probability of the model for the token $y_t$ given the input sequence $x$ and the tokens previously generated, $y_{<t}$.

\begin{equation}
    \label{eq:loss-generative}
    L_{G} = \displaystyle\sum_{t=1}^T \log P(y_t | y_{<t}, x)
\end{equation}

Following the answer generation, the model performs the binary classification on the decoder output (\ie generated answer). We employed the binary cross-entropy loss, denoted as $L_{BCE}$ in Equation~\ref{eq:bce-loss}, which measures the difference between the true labels $y_i$ and the predicted probabilities $\sigma(\hat{y}_i)$, where $\sigma$ denotes the sigmoid function that maps the output of the classifier, $\hat{y}_i$, to a probability between $0$ and $1$. The term $w$ is a weight for the positive class that can handle imbalance by giving more importance to one class over the other.

\begin{equation}
    \label{eq:bce-loss}
    L_{BCE} = - \frac{1}{N} \sum_{i=1}^N [ w y_i \log(\sigma(\hat{y}_i)) + (1 - y_i) \log(1 - \sigma(\hat{y}_i)) ]
\end{equation}

We apply mean pooling on the decoder output and classify it using the same objective function presented in Equation~\ref{eq:bce-loss}. Finally, the overall loss $L$ can be rewritten as $L = L_{G} + L_{BCE}$.

\section{Evaluation}
\label{sec:results}
We evaluate \model using the dataset created and described in Section~\ref{sec:dataset}. 
Our evaluation is divided into MRC and RoB inference. The goal of the former is to identify the sentence that provides evidence that supports the RoB assessment, and the latter is the evaluation of the RoB assessment itself. 

We start by presenting the setup for the \model models, then we present our baselines and evaluation metrics. Finally, 
we present quantitative and qualitative results, showing the strengths and limitations of our approach.

\subsection{\model Setup}
\label{subsec:experimental-setup}
\extrobin was fine-tuned from the checkpoints of Biomed RoBERTa~\cite{Gururangan+2020} and \genrobin was fine-tuned from the checkpoints of BioBART~\cite{Yuan+2022}, for $10$ epochs using AdamW optimizer~\cite{Loshchilov&Hutter2017}. Biomed RoBERTa and BioBART were chosen based on their demonstrated efficacy in processing biomedical texts and their ability to capture relationships within such data.

For \extrobin, we set the batch size to $8$ with gradient accumulation during $4$ steps and the maximum sequence length to $512$. To mitigate overfitting, we set weight decay as an L2 regularization factor with a coefficient of $0.01$ and L1 regularization of $0.1$. All models were trained with a learning rate ranging between $1e-5$ and $5e-5$. We adopted a cosine decay approach for learning rate adjustment throughout the training phases. The same parameters were used for both MRC and RoB Inference.
In \genrobin, we set the batch size and gradient accumulation to $4$ and weight decay and L1 regularization to $0.1$.

\subsection{Baselines}
\label{sec:baselines}
The following baselines were used in our experiments.

\paragraph{\textbf{LLM}}
Although it is not the focus of the present work to extensively evaluate the use of LLMs, we employed them for both MRC and RoB classification tasks. 
Our analysis relied on closed and open models, namely GPT-4o-mini 8B~\cite{OpenAI2024}, Llama3.1 8B~\cite{Dubey+2024}, and Gemma2 9B~\cite{GemmaTeam2024}. 
We created prompts for 0-shot, 1-shot, and few-shot learning scenarios. 
For each instance in the dataset, the LLM received a system role prompt and an instance for the user role. The system role prompt instructed the model to answer RoB classification questions based on the provided context. The possible answers were \textit{yes} and \textit{no}, corresponding to the RoB labels \textit{low} and \textit{high}/\textit{unclear}, respectively.

The specific prompt provided to the models was as follows:

\footnotesize
\begin{verbatim}
You are a specialist performing risk of bias assessments in biomedical articles.
Your input is a question and a context. Your task is to return a JSON document containing 
the answer to the question and an excerpt from the context containing the evidence
that justifies the answer. The possible answers are YES and NO. If there is evidence, 
the output should be in the following 
    format: 
    {
      'answer': '<YES|NO>', 
      'evidence': '<evidence>'
    } 
    If there is no evidence, your answer should be in the following format:
    {
      'answer': 'NO', 
      'evidence': ''
    }.
\end{verbatim}
\normalsize

In the case of 1-shot learning, an additional instance was sampled from the training set and presented to both the user and assistant roles to guide the model. Three instances were sampled from the training set for few-shot learning. This setup assesses how well the models could generalize and accurately classify the RoB based on examples provided within the prompt.

\paragraph{\textbf{Traditional Machine Learning Algorithms}} We also implemented baselines using traditional Machine Learning algorithms such as SVM and LR since they have been extensively used for RoB inference~\cite{Marshall+2015, Marshall+2016,  Pereira+2020, Millard+2015, Marshall+2020}. 
To create the instances, we concatenated the question and the expected answer. Then, we computed the TF-IDF matrix with the 1K most frequent [1,2,3]-grams in the corpus. 

\subsection{Evaluation Metrics}
\label{sec:metrics}
To evaluate the MRC task, we used the traditional metrics F1 score and Exact Match (EM). The F1 score considers partial matches and is computed based on the overlap of tokens between the predicted answer and the ground truth answer. 
EM is more strict since it requires that the predicted output exactly matches the ground truth. 
We also used BERTScore~\cite{Zhang+2020}, which is a language generation evaluation metric based on contextual embeddings.
BERTScore computes precision and recall using cosine similarity between the tokens, and then computes the F1-score (denoted $F_{BERT}$) as the harmonic mean of precision and recall.

For RoB inference, a classification label of low RoB is the positive outcome, while the high/unclear RoB is the negative outcome. True positives ($tp$) and true negatives ($tn$) correspond to the correct classification of low and high/unclear RoB, respectively. False positives ($fp$) occur when a high/unclear risk trial is misclassified as low risk, and false negatives ($fn$) happen when a low-risk trial is misclassified as high/unclear risk.
The metrics used to evaluate the RoB inference task were precision, recall, macro F1-score, and AUC ROC.  

We used T-tests to assess whether the differences were statistically significant. Statistical analyses were performed using the Python library \emph{scypi.stats}. All tests with $p < 0.05$ were considered statistically significant.

\subsection{MRC Results}
\label{sec:MRC-results}
This section presents the results of our experiments in the MRC task compared with LLM baselines. Results for the \model models are the average of ten runs. However, due to their cost, experiments with LLMs were run just once for the entire test set.

Table~\ref{tab:mrc-benchmark} shows the MRC results for each model by bias type. 
We also include the overall performance. Note that the overall scores are calculated for the entire dataset regardless of the bias type. Since some bias types are more prevalent than others, the overall score is not the same as averaging the results for the bias types.

\extrobin outperforms all models in both F1, $F_{BERT}$, and EM for all bias types.
Some errors in \extrobin refer to when the sentence should end. For example, in some results, the sentence ended early at the first punctuation mark.

\begin{table}[tbh]
\caption{Model Comparison in the MRC Task. (Best results in red)}
\label{tab:mrc-benchmark}
\resizebox{\columnwidth}{!}{%
\begin{tabular}{c|c|c|c|ccc|ccc|ccc}
\toprule
\multicolumn{1}{c|}{\multirow{2}{*}{\textbf{\begin{tabular}[c]{@{}c@{}}\textbf{Bias Type}\end{tabular}}}} & \multicolumn{1}{|c|}{\multirow{2}{*}{\textbf{Metric}}} & \multicolumn{1}{c|}{\multirow{2}{*}{\textbf{\genrobin}}} & \multicolumn{1}{l|}{\multirow{2}{*}{\textbf{\extrobin}}} & \multicolumn{3}{c|}{\textbf{Llamma 3.1 8B}}                                                        & \multicolumn{3}{c|}{\textbf{Gemma2 9B}}                                                            & \multicolumn{3}{c}{\textbf{GPT-4o-Mini 8B}}                                                       \\ \cline{5-13} 
\multicolumn{1}{c|}{}                                                                              & \multicolumn{1}{c|}{}                                 & \multicolumn{1}{c|}{}                                          & \multicolumn{1}{l|}{}                                          & \multicolumn{1}{l}{0-shot} & \multicolumn{1}{l}{1-shot} & \multicolumn{1}{l|}{few-shot} & \multicolumn{1}{l}{0-shot} & \multicolumn{1}{l}{1-shot} & \multicolumn{1}{l|}{few-shot} & \multicolumn{1}{l}{0-shot} & \multicolumn{1}{l}{1-shot} & \multicolumn{1}{l}{few-shot} \\ \hline
\multirow{3}{*}{\textbf{Overall}}                                                                   & \textbf{F1}                                           & 93.81                                                          & \red{97.10}                                                 & 40.95                       & 50.99                       & 53.25                         & 50.78                       & 58.74                       & 61.90                         & 55.54                       & 52.47                       & 56.71                         \\
                                                                            & \textbf{$F_{BERT}$}                                   & 89.52                                                          & \red{91.30}                                                 & 58.97                       & 70.30                       & 74.68                         & 64.36                       & 76.80                       & 80.91                         & 72.85                       & 68.97                       & 75.36                         \\
                                                                            & \textbf{EM}                                           & 82.09                                                          & \red{87.76}                                                 & 17.51                       & 24.94                       & 27.29                         & 28.64                       & 33.84                       & 34.99                         & 29.76                       & 29.99                       & 32.37                         \\\midrule
                                                                            
\multirow{3}{*}{\textbf{Detection}}                                                                 & \textbf{F1}                                           & 95.55                                                          & \red{97.66}                                                 & 28.14                       & 55.31                       & 60.33                         & 60.06                       & 66.28                       & 69.27                         & 61.92                       & 57.87                       & 62.55                         \\
                                                                            & \textbf{$F_{BERT}$}                                   & 99.20                                                          & \red{99.47}                                                 & 41.65                       & 74.30                       & 83.30                         & 79.06                       & 88.06                       & 91.12                         & 83.09                       & 76.55                       & 83.56                         \\
                                                                            & \textbf{EM}                                           & 87.90                                                          & \red{91.54}                                                 & 16.88                       & 35.21                       & 39.17                         & 42.29                       & 46.04                       & 48.96                         & 39.79                       & 39.79                       & 41.25                         \\\midrule
\multirow{3}{*}{\textbf{Performance}}                                                               & \textbf{F1}                                           & 94.73                                                          & \red{97.76}                                                 & 34.70                       & 56.85                       & 60.07                         & 60.94                       & 65.86                       & 69.46                         & 62.19                       & 61.79                       & 63.89                         \\
                                                                            & \textbf{$F_{BERT}$}                                   & 99.00                                                          & \red{99.32}                                                 & 52.00                       & 78.54                       & 86.13                         & 81.09                       & 88.66                       & 91.97                         & 86.53                       & 84.16                       & 87.92                         \\
                                                                            & \textbf{EM}                                           & 85.74                                                          & \red{91.95}                                                 & 18.83                       & 34.63                       & 38.53                         & 41.34                       & 44.59                       & 47.62                         & 38.96                       & 41.56                       & 42.86                         \\\midrule
\multirow{3}{*}{\textbf{Selection}}                                                                 & \textbf{F1}                                           & 94.86                                                          & \red{97.33}                                                 & 54.81                       & 57.03                       & 58.30                         & 57.61                       & 63.65                       & 65.51                         & 60.94                       & 57.16                       & 60.10                         \\
                                                                            & \textbf{$F_{BERT}$}                                   & 99.01                                                          & \red{99.23}                                                 & 80.93                       & 82.25                       & 86.20                         & 80.44                       & 88.71                       & 91.31                         & 86.48                       & 80.63                       & 84.05                         \\
                                                                            & \textbf{EM}                                           & 83.01                                                          & \red{88.06}                                                 & 22.75                       & 26.30                       & 27.53                         & 30.07                       & 35.75                       & 35.42                         & 31.25                       & 30.63                       & 33.28                         \\\midrule
\multirow{3}{*}{\textbf{Attrition}}                                                                 & \textbf{F1}                                           & 89.80                                                          & \red{97.49}                                                 & 16.22                       & 22.94                       & 36.64                         & 35.26                       & 42.82                       & 46.29                         & 45.93                       & 40.19                       & 40.35                         \\
                                                                            & \textbf{$F_{BERT}$}                                   & 97.99                                                          & \red{98.88}                                                 & 35.34                       & 48.69                       & 70.78                         & 60.68                       & 75.24                       & 82.67                         & 74.94                       & 67.91                       & 72.08                         \\
                                                                            & \textbf{EM}                                           & 75.05                                                          & \red{81.57}                                                 & 2.78                        & 2.78                        & 11.11                         & 11.11                       & 15.74                       & 16.67                         & 15.74                       & 18.52                       & 17.13                         \\\midrule
\multirow{3}{*}{\textbf{Reporting}}                                                                 & \textbf{F1}                                           & 88.38                                                          & \red{92.85}                                                 & 16.50                       & 27.73                       & 25.59                         & 9.30                        & 30.70                       & 39.13                         & 11.29                       & 14.17                       & 29.96                         \\
                                                                            & \textbf{$F_{BERT}$}                                   & 97.74                                                          & \red{98.38}                                                 & 50.69                       & 70.57                       & 58.57                         & 21.69                       & 67.43                       & 85.33                         & 21.15                       & 29.45                       & 71.47                         \\
                                                                            & \textbf{EM}                                           & 67.74                                                          & \red{78.33}                                                 & 4.17                        & 11.90                       & 10.71                         & 3.57                        & 12.50                       & 14.29                         & 6.55                        & 5.95                        & 11.90                         \\\midrule
\multirow{3}{*}{\textbf{Other}}                                                                     & \textbf{F1}                                           & 89.35                                                          & \red{95.94}                                                 & 20.62                       & 32.67                       & 29.09                         & 14.48                       & 33.93                       & 41.27                         & 34.92                       & 32.17                       & 43.14                         \\
                                                                            & \textbf{$F_{BERT}$}                                   & 97.88                                                          & \red{98.65}                                                 & 54.64                       & 79.75                       & 74.14                         & 34.48                       & 77.41                       & 88.99                         & 74.00                       & 73.44                       & 88.90                         \\
                                                                            & \textbf{EM}                                           & 74.90                                                          & \red{83.24}                                                 & 3.64                        & 8.94                        & 10.60                         & 5.63                        & 11.59                       & 15.56                         & 13.91                       & 14.57                       & 19.21                        
\\\bottomrule
\end{tabular}}
\end{table}

\genrobin performs slightly lower than \extrobin but still better than the LLMs. 
Some of the errors made by \genrobin refer to word replacement. 
For example, the ground truth contains ``\emph{\textbf{simultaneously}, the radiological evaluation is reviewed by 3 independent...}'', and in the generated answer has ``\emph{\textbf{therefore}, the radiological evaluation is reviewed by 3 independent...}''.
Also, there are cases in which the generated answers are incorrect pieces of the context. Most errors in the exact match performance occur in the Reporting Bias. 

Looking at the predictions generated by the LLMs, we noticed that some of the responses tend to be longer and include more details than the ground truth. This ends up lowering the scores significantly because of the nature of the metrics.

There are also cases where the LLM gives the correct evidence of an additional one. For example, a given instance of the dataset has the following question-answer tuple: ``\emph{Was the allocation sequence random?} - \emph{randomization was stratified by patients' BMI (non-obese $<$30 kg/m2 and obese 30-35 kg/m2) using mixed block sizes}''. 
Both \extrobin and \genrobin generated an exact match. 
Llama3.1 retrieved the answer with additional evidence (\eg ``\emph{randomization was stratified by patients' BMI (non-obese $<$30 kg/m2 and obese 30-35 kg/m2) using mixed block sizes. ... randomization list created by the trial statistician using nquery advisor v6.0 software.}''). GPT-4o-Mini and Gemma2 generate abstractive content based on the evidence. Llamma3.1 has the poorest performance in the 0-shot scenario, and all the LLMs struggle to retrieve evidence in the attrition bias, other bias, and reporting bias types.

Among the LLMs, Gemma2 in the few-shot scenario has the highest overall F1 score, while Llamma had the lowest.
We speculate that the superiority is due to Gemma being the largest among the three.

\subsection{RoB Inference Results}
\label{sec:robin-results}
Table~\ref{tab:results-by-type-of-bias} presents the results for the RoB inference task. Notice that the scores are macro-averaged, so the values for F1 are not exactly the harmonic mean between precision and recall.
Overall, \extrobin was the best performer in terms of F1 and precision, and \genrobin achieved the best recall. 
A statistical test showed that the differences between \extrobin and \genrobin are not significant. In comparison with the baselines, the differences are indeed significant.
The worst model overall was LR.
Among the generative models, Gemma was again the best performer, repeating the results in the previous 

\extrobin has shown to be a more robust option for the RoB classification task due to its performance in the different bias types -- it is the top scorer in terms of F1 in 4 out of 6 bias types (Detection, Selection, Reporting, and other). 

In performance bias, GPT achieved the best results in all three metrics, followed by the other LLMs. Our models were able to outperform only the classical algorithms.

In attrition bias, again, we have an LLM as the best performer in terms of F1 -- Llamma had the best F1, followed by \genrobin, which had the best precision and recall. Attrition bias is one of the most difficult bias types to assess, as it usually relies on checking for withdrawals and dropouts in tables, and in many cases, our automatic annotation process was unable to retrieve the tables needed for such an assessment.

\begin{table}[]
\caption{Results for the risk of bias inference task by type of bias. Best results in red.}
\label{tab:results-by-type-of-bias}
\resizebox{\columnwidth}{!}{%
\begin{tabular}{@{}c|c|c|c|c|c|ccc|ccc|ccc@{}}
\toprule
\multirow{2}{*}{\textbf{\begin{tabular}[c]{@{}c@{}}Bias \\ Type\end{tabular}}} & \multicolumn{1}{l|}{\multirow{2}{*}{\textbf{Metric}}} & \multicolumn{1}{l|}{\multirow{2}{*}{\textbf{\genrobin}}} & \multicolumn{1}{l|}{\multirow{2}{*}{\textbf{\extrobin}}} & \multirow{2}{*}{\textbf{SVM}} & \multirow{2}{*}{\textbf{LR}} & \multicolumn{3}{c|}{\textbf{Llamma3.1 8B}}                                                & \multicolumn{3}{c|}{\textbf{Gemma2 9B}}                                                   & \multicolumn{3}{c}{\textbf{GPT-4o-Mini 8B}} 
\\ \cline{7-15}
                                                                               & \multicolumn{1}{l|}{}                                 & \multicolumn{1}{l|}{}                                                   & \multicolumn{1}{l|}{}                                                   &                               &                              & \multicolumn{1}{l}{0-shot} & \multicolumn{1}{l}{1-shot} & \multicolumn{1}{l|}{few-shot} & \multicolumn{1}{l}{0-shot} & \multicolumn{1}{l}{1-shot} & \multicolumn{1}{l|}{few-shot} & \multicolumn{1}{c}{0-shot}         & \multicolumn{1}{c}{1-shot} & few-shot       \\
\hline
\multirow{3}{*}{\textbf{Overall}}                                              & \textbf{F1}                                           & 72.75                                                                   & \red{74.18}                                                                   & 71.29                         & 69.11                        & \multicolumn{1}{c}{64.54}  & \multicolumn{1}{c}{68.15}  & 68.12                         & \multicolumn{1}{c}{67.57}  & \multicolumn{1}{c}{67.56}  & 70.03                         & \multicolumn{1}{c}{69.20}          & \multicolumn{1}{c}{69.62}  & 69.60          \\
                                                                               & \textbf{Prec.}                                        & 73.21                                                                   & \red{75.49}                                                          & 70.84                         & 68.96                        & \multicolumn{1}{c}{65.06}  & \multicolumn{1}{c}{67.75}  & 68.43                         & \multicolumn{1}{c}{67.18}  & \multicolumn{1}{c}{68.04}  & 71.16                         & \multicolumn{1}{c}{68.83}          & \multicolumn{1}{c}{69.17}  & 69.14          \\
                                                                               & \textbf{Rec.}                                         & \red{74.51}                                                          & 73.62                                                                   & 73.02                         & 71.26                        & \multicolumn{1}{c}{67.15}  & \multicolumn{1}{c}{69.35}  & 67.87                         & \multicolumn{1}{c}{68.39}  & \multicolumn{1}{c}{67.21}  & 69.35                         & \multicolumn{1}{c}{69.78}          & \multicolumn{1}{c}{70.99}  & 70.72          \\
\hline
\multirow{3}{*}{\textbf{Detection}}                                            & \textbf{F1}                                           & 69.56                                                                   & \red{72.56}                                                          & 62.19                         & 59.57                        & \multicolumn{1}{c}{45.21}  & \multicolumn{1}{c}{61.45}  & 64.17                         & \multicolumn{1}{c}{65.22}  & \multicolumn{1}{c}{69.22}  & 68.66                         & \multicolumn{1}{c}{64.78}          & \multicolumn{1}{c}{64.44}  & 66.12          \\
                                                                               & \textbf{Prec.}                                        & 70.75                                                                   & \red{74.39}                                                          & 63.50                         & 59.40                        & \multicolumn{1}{c}{57.92}  & \multicolumn{1}{c}{61.74}  & 63.64                         & \multicolumn{1}{c}{64.70}  & \multicolumn{1}{c}{68.92}  & 69.03                         & \multicolumn{1}{c}{64.27}          & \multicolumn{1}{c}{64.26}  & 65.49          \\
                                                                               & \textbf{Rec.}                                         & 70.20                                                                   & \red{71.76}                                                          & 61.69                         & 60.03                        & \multicolumn{1}{c}{57.96}  & \multicolumn{1}{c}{64.64}  & 65.38                         & \multicolumn{1}{c}{67.50}  & \multicolumn{1}{c}{69.56}  & 68.34                         & \multicolumn{1}{c}{66.86}          & \multicolumn{1}{c}{67.52}  & 68.21          \\
\hline
\multirow{3}{*}{\textbf{Performance}}                                          & \textbf{F1}                                           & 72.17                                                                   & 72.91                                                                   & 64.85                         & 63.14                        & \multicolumn{1}{c}{54.33}  & \multicolumn{1}{c}{76.25}  & 75.19                         & \multicolumn{1}{c}{75.23}  & \multicolumn{1}{c}{74.25}  & 75.28                         & \multicolumn{1}{c}{76.37}          & \multicolumn{1}{c}{75.75}  & \red{76.40} \\
                                                                               & \textbf{Prec.}                                        & 74.07                                                                   & 74.14                                                                   & 69.55                         & 68.06                        & \multicolumn{1}{c}{66.75}  & \multicolumn{1}{c}{76.20}  & 75.81                         & \multicolumn{1}{c}{75.23}  & \multicolumn{1}{c}{74.21}  & 75.68                         & \multicolumn{1}{c}{76.50}          & \multicolumn{1}{c}{76.19}  & \red{76.81} \\
                                                                               & \textbf{Rec.}                                         & 73.16                                                                   & 72.88                                                                   & 67.40                         & 65.87                        & \multicolumn{1}{c}{60.52}  & \multicolumn{1}{c}{76.37}  & 74.99                         & \multicolumn{1}{c}{75.47}  & \multicolumn{1}{c}{74.29}  & 75.12                         & \multicolumn{1}{c}{76.76}          & \multicolumn{1}{c}{76.34}  & \red{76.98} \\
\hline
\multirow{3}{*}{\textbf{Selection}}                                            & \textbf{F1}                                           & 71.49                                                                   & \red{72.13}                                                                   & 71.32                         & 68.72                        & \multicolumn{1}{c}{62.28}  & \multicolumn{1}{c}{62.00}  & 61.82                         & \multicolumn{1}{c}{59.95}  & \multicolumn{1}{c}{60.70}  & 60.92                         & \multicolumn{1}{c}{62.14}          & \multicolumn{1}{c}{65.17}  & 64.96          \\
                                                                               & \textbf{Prec.}                                        & 71.63                                                                   & \red{76.31}                                                                   & 71.50                         & 68.00                        & \multicolumn{1}{c}{63.51}  & \multicolumn{1}{c}{61.90}  & 62.96                         & \multicolumn{1}{c}{60.09}  & \multicolumn{1}{c}{63.98}  & 65.07                         & \multicolumn{1}{c}{62.75}          & \multicolumn{1}{c}{64.72}  & 64.39          \\
                                                                               & \textbf{Rec.}                                         & \red{74.03}                                                                   & 69.97                                                                   & 71.18                         & 69.69                        & \multicolumn{1}{c}{61.52}  & \multicolumn{1}{c}{62.10}  & 61.11                         & \multicolumn{1}{c}{59.83}  & \multicolumn{1}{c}{59.59}  & 59.69                         & \multicolumn{1}{c}{61.68}          & \multicolumn{1}{c}{65.73}  & 65.75          \\
\hline
\multirow{3}{*}{\textbf{Attrition}}                                            & \textbf{F1}                                           & 57.11                                                                   & 56.00                                                                   & 45.26                         & 47.64                        & \multicolumn{1}{c}{43.03}  & \multicolumn{1}{c}{46.15}  & \red{57.81}                & \multicolumn{1}{c}{48.75}  & \multicolumn{1}{c}{47.29}  & 51.36                         & \multicolumn{1}{c}{43.57}          & \multicolumn{1}{c}{49.13}  & 48.22          \\
                                                                               & \textbf{Prec.}                                        & \red{60.44}                                                          & 59.19                                                                   & 55.27                         & 52.24                        & \multicolumn{1}{c}{47.29}  & \multicolumn{1}{c}{49.35}  & 57.83                         & \multicolumn{1}{c}{49.04}  & \multicolumn{1}{c}{47.65}  & 52.19                         & \multicolumn{1}{c}{45.04}          & \multicolumn{1}{c}{49.47}  & 48.36          \\
                                                                               & \textbf{Rec.}                                         & \red{59.49}                                                          & 58.02                                                                   & 53.12                         & 51.91                        & \multicolumn{1}{c}{47.83}  & \multicolumn{1}{c}{49.42}  & 57.96                         & \multicolumn{1}{c}{49.11}  & \multicolumn{1}{c}{47.88}  & 51.93                         & \multicolumn{1}{c}{46.60}          & \multicolumn{1}{c}{49.51}  & 48.44          \\
\hline
\multirow{3}{*}{\textbf{Reporting}}                                            & \textbf{F1}                                           & 55.56                                                                   & \red{62.40}                                                          & 45.98                         & 49.20                        & \multicolumn{1}{c}{45.28}  & \multicolumn{1}{c}{51.33}  & 52.35                         & \multicolumn{1}{c}{42.03}  & \multicolumn{1}{c}{43.21}  & 58.00                         & \multicolumn{1}{c}{40.29}          & \multicolumn{1}{c}{38.73}  & 47.29          \\
                                                                               & \textbf{Prec.}                                        & 61.02                                                                   & \red{63.70}                                                          & 61.21                         & 62.69                        & \multicolumn{1}{c}{50.36}  & \multicolumn{1}{c}{56.19}  & 52.53                         & \multicolumn{1}{c}{49.22}  & \multicolumn{1}{c}{43.84}  & 59.09                         & \multicolumn{1}{c}{52.22}          & \multicolumn{1}{c}{46.90}  & 51.65          \\
                                                                               & \textbf{Rec.}                                         & 57.89                                                                   & \red{62.79}                                                          & 53.87                         & 55.47                        & \multicolumn{1}{c}{50.20}  & \multicolumn{1}{c}{54.20}  & 52.55                         & \multicolumn{1}{c}{49.70}  & \multicolumn{1}{c}{44.76}  & 58.34                         & \multicolumn{1}{c}{50.50}          & \multicolumn{1}{c}{49.30}  & 51.05          \\
\hline
\multirow{3}{*}{\textbf{Other}}                                                & \textbf{F1}                                           & 60.89                                                                   & \red{67.03}                                                          & 49.63                         & 45.08                        & \multicolumn{1}{c}{54.35}  & \multicolumn{1}{c}{59.55}  & 62.84                         & \multicolumn{1}{c}{44.25}  & \multicolumn{1}{c}{46.99}  & 64.55                         & \multicolumn{1}{c}{58.39}          & \multicolumn{1}{c}{54.59}  & 60.25          \\
                                                                               & \textbf{Prec.}                                        & 65.83                                                                   & 68.65                                                                   & 63.68                         & 58.94                        & \multicolumn{1}{c}{67.58}  & \multicolumn{1}{c}{67.21}  & 63.08                         & \multicolumn{1}{c}{50.33}  & \multicolumn{1}{c}{47.36}  & 67.57                         & \multicolumn{1}{c}{\red{75.00}} & \multicolumn{1}{c}{68.99}  & 67.84          \\
                                                                               & \textbf{Rec.}                                         & 61.54                                                                   & \red{67.12}                                                          & 53.44                         & 51.45                        & \multicolumn{1}{c}{56.11}  & \multicolumn{1}{c}{59.40}  & 62.65                         & \multicolumn{1}{c}{50.07}  & \multicolumn{1}{c}{48.05}  & 63.62                         & \multicolumn{1}{c}{59.02}          & \multicolumn{1}{c}{56.35}  & 59.94         \\
                                                                               \bottomrule
\end{tabular}}
\end{table}

\begin{figure*}[ht!]
    \centering
    \begin{subfigure}[b]{0.49\textwidth}
        \centering        \includegraphics[width=\linewidth]{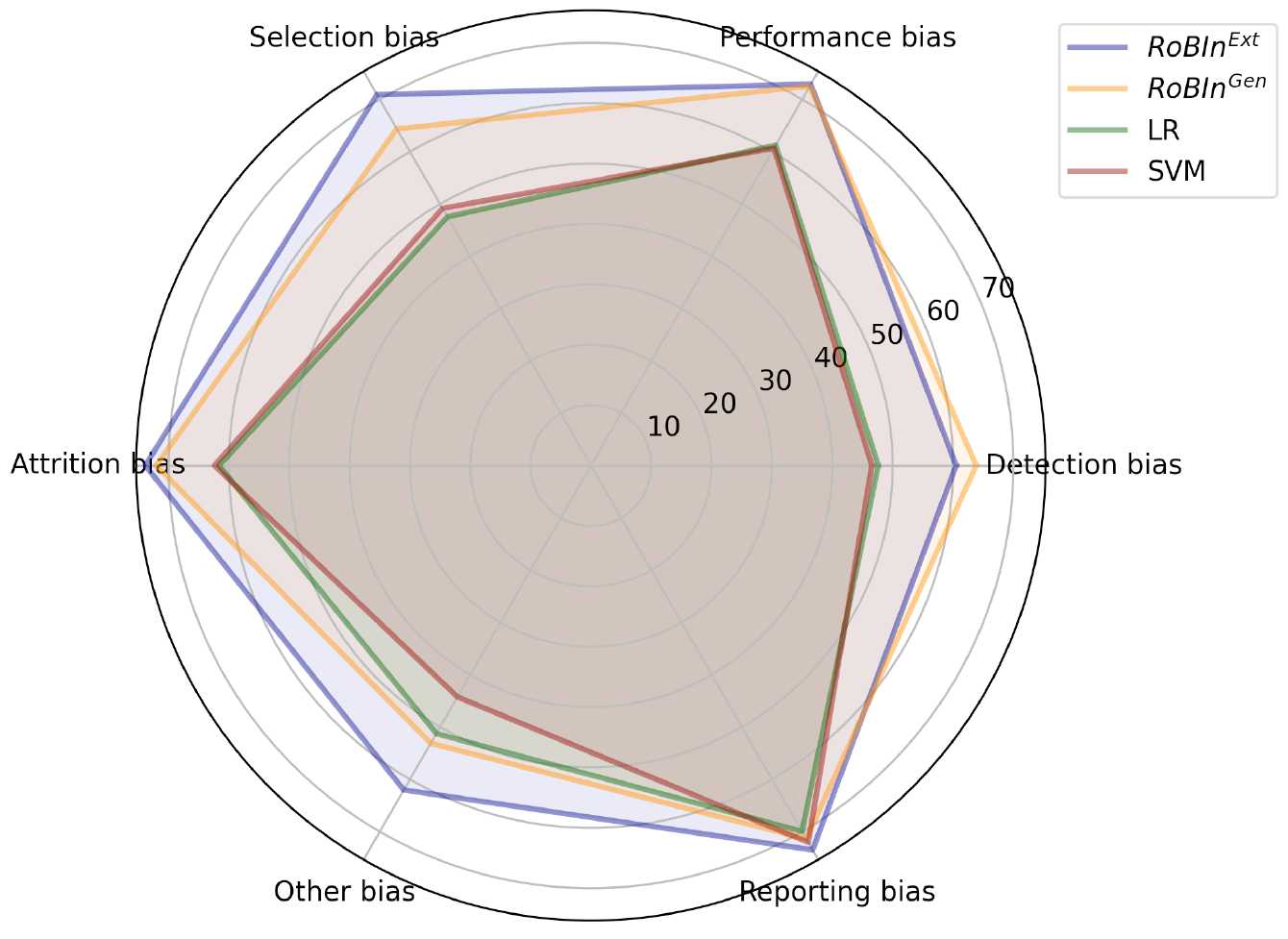}
        \caption{Macro F1}
        \label{fig:f1-type-bias}
    \end{subfigure}
    \begin{subfigure}[b]{0.49\linewidth}
        \centering        
\includegraphics[width=\textwidth]{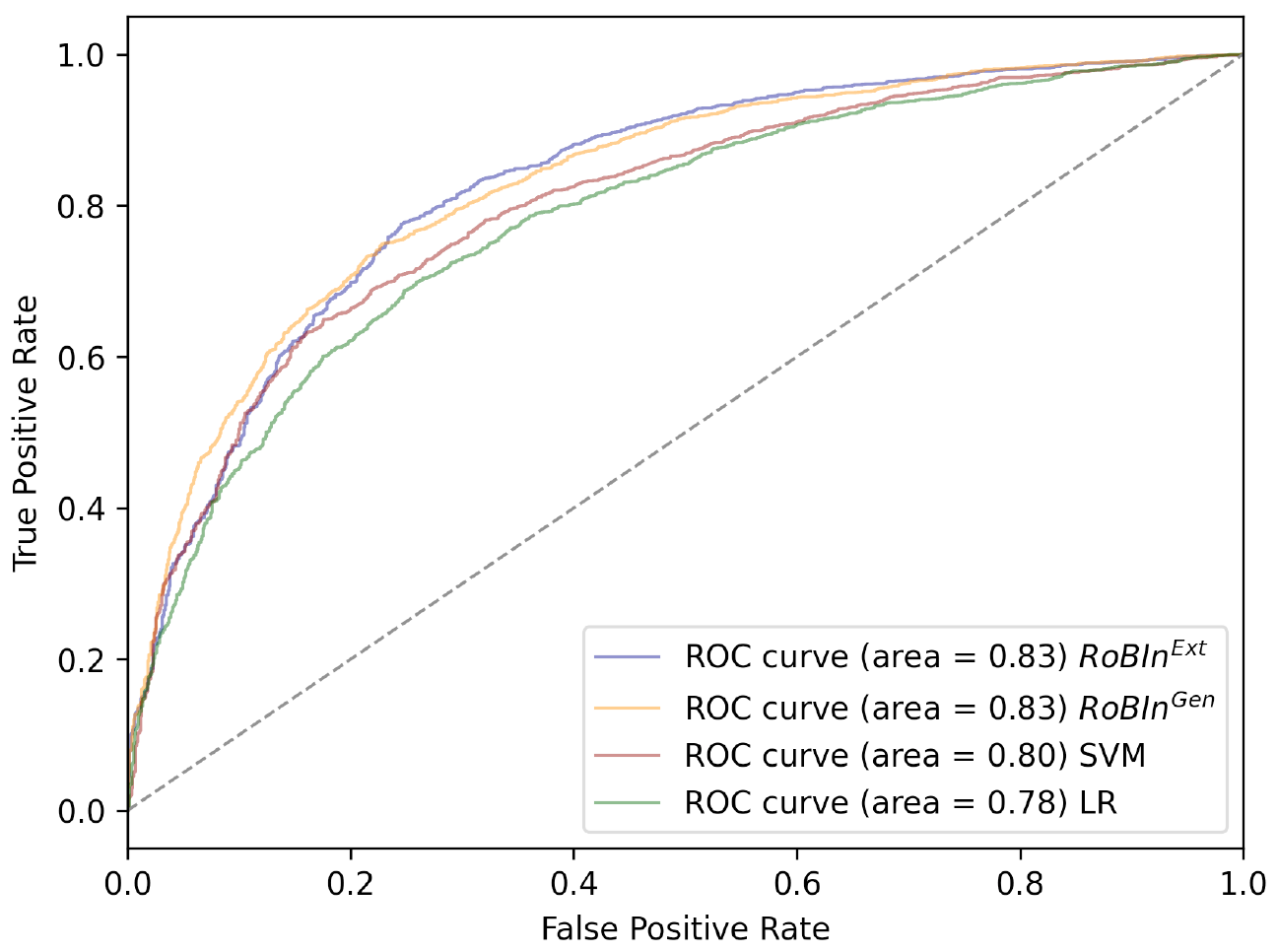}
        \caption{ROC AUC Curve of the best models.}
        \label{fig:results-roc-curve}
    \end{subfigure}    
    \caption{Macro F1 and ROC AUC Curves.}
    \label{fig:radar-plot-roc-curve}
\end{figure*}

Figure~\ref{fig:radar-plot-roc-curve} depicts the performance of the models in terms of F1 score and AUC ROC, with a specific focus on bias types and overall performance. Figure~\ref{fig:f1-type-bias} compares the F1 score of the models. The overall performance appears similar, with minor variations across the different models. \extrobin shows slightly better performance than the other models, indicating their robustness in scenarios with reporting, detection, and other biases. \genrobin shows consistent performance with a decrease in the other bias and slightly better in the attrition. SVM and LR models show a dip in performance compared to the \model models, but SVM performs better than LR in selection, detection, and other biases.

The radar plot (Figure~\ref{fig:f1-type-bias}) shows that the \model models have strong and consistent performance across different bias types and overall metrics. This suggests these models may be more reliable and robust in the RoB inference task.

Figure~\ref{fig:results-roc-curve} shows that \extrobin and \genrobin have the highest ROC AUC ($0.83$), indicating they perform well overall and are good at distinguishing between classes. SVM has a moderate F1 and ROC AUC $0.80$, performing well but slightly less consistently across different thresholds than \model. LR also shows a slight drop from F1 and ROC AUC (\ie $0.78$), indicating some threshold-specific performance variance. 

We noticed that \model correctly classified between $77\%$ and $93\%$ of the positive class instances (\ie instances with low RoB), except for the other bias and reporting bias categories, where the performance decreases for the positive instance classification. 
All these findings are illustrated in the confusion matrices shown in the supplementary material.

For the negative class, \model models struggle to classify them in the detection, attrition, and selection bias categories. \genrobin shows more balanced results in the performance bias with almost $78\%$ of the positive instances correctly classified and $75\%$ of the negative instances. On the other hand, \genrobin showed the lowest performance in the negative class in the detection and selection bias categories, incurring more false positives (\ie instances of high RoB being classified as low RoB).

In general, the LR model has the poorest performance concerning the positive class with results below $50\%$, except in the selection and detection bias categories. However, it showed good results in classifying the negative class for most bias types. SVM presents results similar to those of the LR model.

In summary, \genrobin is robust in classifying positive instances but less reliable in negative instances. \extrobin performs well and is consistent and has a slightly better performance compared to \genrobin. SVM and LR have the poorest performance compared to \model models. The results in the LR and SVM indicate that such models might be biased to the negative class for some bias types.

Finally, when comparing our results with the scores obtained in related works (see Table~\ref{tab:comparison-related-work}), we find scores within the same range. Nevertheless, direct comparisons cannot be performed since the datasets are different.

\section{Conclusion}
\label{sec:conclusion}
We introduced the \model dataset, a dataset automatically built for the tasks of MRC and RoB inference. The dataset is structured as an MRC dataset with binary classification. Additionally, we have introduced \model, a two-step model for retrieving evidence from clinical trial publications and performing RoB inference. We proposed a novel model that outperforms the state-of-the-art approaches and LLMs in many scenarios. 
Our model identifies snippets of text that can be used as a supporting sentence to perform RoB judgments. Both \model models are robust in the MRC task, with \extrobin being more efficient. 

Based on the extracted/generated evidence, \model performs binary classification to decide whether the trial is at a \textit{low} RoB or a \textit{high}/\textit{unclear} RoB. We show that both \genrobin and \extrobin are robust and have the best discriminatory capability with ROC AUC $0.83$. 

The \model dataset encapsulates an essential portion of the biases identified within the dataset sourced from the Cochrane collaboration. However, it does so on a significantly smaller scale, attributed to the automated acquisition process of the dataset. While maintaining a representation of diverse biases, this size reduction and imbalance presents unique challenges, particularly in MRC scenarios. Working with such a compact dataset demands models to recognize and accurately extract relevant information about biases from a constrained set of examples.

Another limitation is that our dataset was automatically annotated from systematic reviews and with little human intervention. The instances were validated by a computer scientist who had little experience with systematic reviews before but had reasonable knowledge of the Cochrane Risk of Bias tool. During validation, some instances were discarded because they mentioned existing details in the clinical trial protocol, and our approach uses only clinical trial publications. The dataset also presents an imbalance with respect to the classes, with more trials judged as a low RoB than the opposite. The imbalance of the dataset is a harmful factor and a challenge to solve since most errors occur when predicting the negative class, which is less represented in the dataset. The imbalance also happens in the bias types, with more instances in the selection bias category.

A false positive could represent a problematic scenario in MRC for healthcare and the biomedical area. This means that a study with a high or unclear RoB is incorrectly identified as having a low RoB. Such errors could lead to the inclusion of biased evidence in systematic reviews or clinical guidelines, potentially impacting patient care and outcomes.

A possible route to improve performance and decrease the false positive rate is to augment the number of instances in each class by generating synthetic instances to create a balanced dataset. Another possibility is to downsample the selection bias instances. Note that selection bias contains two signaling questions, one related to sequence generation and another concerning allocation concealment, which justifies the more significant number of instances of this bias type.

\section*{Acknowledgments}
 The authors are thankful to Cochrane for giving us access to CDSR and for allowing us to share the dataset that we derived from it.
 This work was partially financed by the Coordenação de Aperfeiçoamento de Pessoal de Nível Superior - Brasil (CAPES) - Finance Code 001 and CNPq.

\bibliographystyle{elsarticle-num-names} 
\bibliography{paper}

\end{document}